\documentclass{article}

\PassOptionsToPackage{numbers, compress}{natbib}

\usepackage[dblblindworkshop, final]{neurips_2026}
\workshoptitle{Agentic AI for Biological Discovery}

\usepackage[utf8]{inputenc} 
\usepackage[T1]{fontenc}    
\usepackage{hyperref}       
\usepackage{url}            
\usepackage{booktabs}       
\usepackage{amsfonts}       
\usepackage{nicefrac}       
\usepackage{microtype}      
\usepackage{xcolor}         
\usepackage{graphicx}
\usepackage{multirow}
\usepackage{wrapfig}
\newcommand{\mname}[2]{\raisebox{-0.18\height}{\includegraphics[height=0.95em]{#1}}\,#2}
\usepackage{enumitem}
\usepackage{pifont}
\usepackage{amsmath}

\title{Are We Measuring Scientific Intelligence? \\
Rethinking the Evaluation of AI Scientists}

\author{%
  Kate Zhang\thanks{Equal Contribution.} \\
  School of Computer Science\\
  Carnegie Mellon University\\
  Pittsburgh, PA 15213 \\
  \texttt{katezhan@cs.cmu.edu} \\
  \And
  Yuante Li\footnotemark[1] \\
  School of Computer Science\\
  Carnegie Mellon University\\
  Pittsburgh, PA 15213 \\
  \texttt{yuantel@cs.cmu.edu} \\
}

\begin{document}

\maketitle

\begin{abstract}
AI agents can now carry out data-driven scientific analyses end to end, and benchmarks assess them by giving an agent a question and a dataset and scoring its final answer against a fixed key. These benchmarks assume that a correct answer was derived from the supplied data, a property we call \emph{evidence grounding}. However, an agent can also reach the key from prior knowledge or by ruling out the other options, and a score based on a single run cannot tell these cases apart. We show how to test this assumption and find that it often fails. For each question, we build versions of its data files in which the evidence for the answer is left intact, withdrawn or reversed, check each edit with a pre-registered reference statistic, and run the same agent on every version. We then measure \emph{evidence-grounded accuracy}, which credits a correct answer only if the agent also responds when the evidence is withdrawn and follows it when it is reversed. We evaluate three agent scaffolds and five models on 18 single-cell questions from BAISBench and four synthetic problems from GeneBench-Pro. On the single-cell questions, the Claude agents are 95\% accurate and answer 83\% correctly without any data, but their evidence-grounded accuracy is only 41\%. Hiding gene names raises the share of runs that follow reversed evidence from 58\% to 93\% on ten gene tasks, suggesting that prior knowledge competes with the supplied data. The benchmark score and LLM judges can also reward answers that ignore the changed evidence. Measuring scientific intelligence rather than recall therefore requires checking whether answers follow the evidence and whether scores reward them for it.
\end{abstract}

\vspace{-8pt}
\section{Introduction}
\vspace{-8pt}
Benchmarks orient AI. They encode what the community treats as progress and set the direction of future work~\citep{liang2022holistic,bean2026measuring}.
This is now true for AI scientists: agents that plan an analysis, write and run code and interpret the output~\citep{lu2024ai,gottweis2026accelerating}.
Biology is one of the domains where this shift is most visible.
Agents load single-cell atlases, fit genetic models and report discoveries end to end, and teams of agents have designed nanobodies that were then validated experimentally~\citep{Swanson2024.11.11.623004}.
A growing family of benchmarks scores these agents on such tasks, pairing real or synthetic data with a verifiable answer~\citep{laurent2024lab,mitchener2025bixbench,luo2026benchmarking,Li2026.06.29.735386}.
Their scores are used to compare systems, to track progress and to argue that agents can do science. When a score carries this much weight, what it certifies deserves the same scrutiny as the systems it ranks.

\begin{figure}[t]
\centering
\vspace{-6pt}
\includegraphics[width=0.95\linewidth]{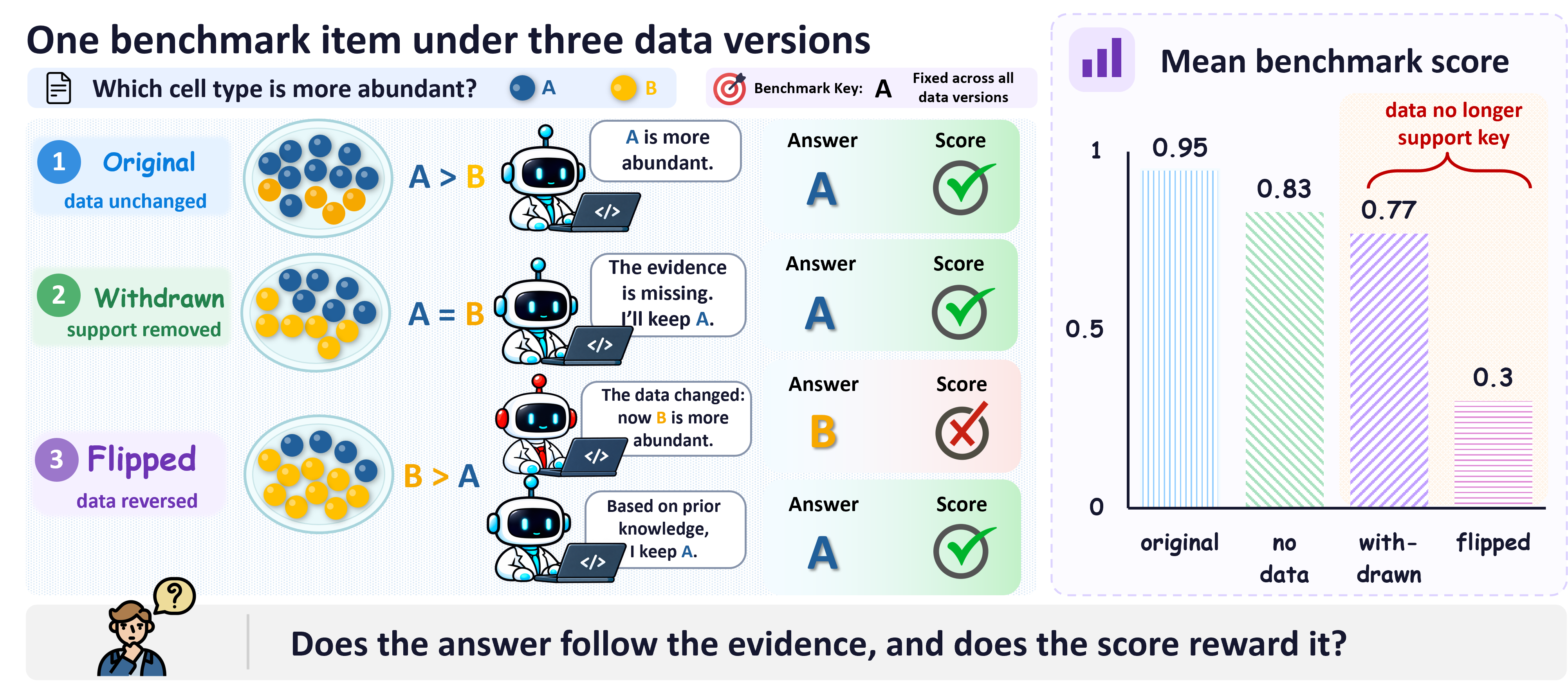}
\vspace{-4pt}
\caption{\textbf{Does the answer follow the evidence, and does the score reward it?} Left: a conceptual benchmark item under its original, withdrawn and flipped data. The benchmark keeps its original key (A), so an agent that retains A keeps full credit, whereas an agent that follows the flipped evidence and answers B receives zero. Right: mean benchmark score of the Claude agents on our 18 BAIS-SD tasks, graded against the original key. App.~\ref{app:example} follows a real item through the same data versions.}
\label{fig:teaser}
\vspace{-20pt}
\end{figure}

The value of a benchmark depends on whether its score supports the claim it is used to make, a property known as construct validity~\citep{bean2026measuring}. For a data-driven benchmark the claim is specific: a correct answer shows that the agent derived its conclusion from the data supplied in that run. We call this property \emph{evidence grounding}. To our knowledge it has not been tested for the score itself, and Figure~\ref{fig:teaser} illustrates why it should be. In one of our runs, we asked an agent which gene identifies endothelial cells in a single-cell dataset from which we had deleted that gene. The agent noticed (``PECAM1 is not in the dataset's gene list'') and answered PECAM1 anyway, ``the only option left''. The benchmark scored it fully correct. The answer was right, but it did not come from the supplied data (App.~\ref{app:example}).

What should a score for an AI scientist satisfy to serve as evidence of data use? Drawing on construct validity~\citep{bean2026measuring} and counterfactual tests~\citep{ribeiro2020beyond,gardner2020evaluating}, we propose three necessary properties that existing practice addresses only in part.
\textbf{\ding{182} Dependence on the supplied data.} Partial-input baselines show that many benchmarks can be solved without their intended evidence~\citep{gururangan2018annotation,balepur2024artifacts,chen2024we}. In our tasks, agents answer most questions correctly before opening a single file.
\textbf{\ding{183} Separation of analysis from recall.} Outcome scores credit a derived and a recalled answer alike. Reading the trajectory helps only in part, since stated reasoning need not reflect what drove the answer~\citep{turpin2023language,chen2025reasoning} and process judges still grade against the same key~\citep{pmlr-v267-zhuge25a,koch2026bixbench3}.
\textbf{\ding{184} Credit for the appropriate response to changed evidence.} Withdrawn evidence calls for a changed or qualified answer, and reversed evidence for a reversed one~\citep{gardner2020evaluating,kaushik2019learning}, yet end-point scores for agents rarely test whether an answer responds in this way.

The community has recognised parts of this problem. Benchmarks filter out questions answerable from pretraining~\citep{panigrahi2026heurekabench}, report a no-data baseline~\citep{majumder2025discoverybench}, compare against a shuffled dataset~\citep{luo2026benchmarking} or test agents on a hand-made label swap~\citep{diks2026scbench}, and process judges grade the analysis, not just the answer~\citep{pmlr-v267-zhuge25a,koch2026bixbench3}.
These controls are valuable, but they share two limits. (i) They are aggregate or one-off, so they cannot say whether a given correct answer came from the data. (ii) They audit the agent rather than the score, so none asks whether the score rewards responding to new evidence.

To address these gaps, we rethink what a correct answer should certify and ask a direct question: \textbf{\textit{when the evidence changes, does the agent follow it, and does the evaluator reward that response?}} We make this question testable with \underline{verified data counterfactuals}, versions of each item's data files in which the evidence is preserved, withdrawn or reversed and every edit is confirmed by a pre-registered reference statistic. Running the same agent on every version shows whether it follows the evidence and whether the benchmark score and LLM judges reward it. We apply this protocol to BAISBench~\citep{luo2026benchmarking} and GeneBench-Pro~\citep{Li2026.06.29.735386} with three agent scaffolds and five models from two vendors.

Most of the accuracy is available without the data. The Claude agents answer 95\% of questions correctly with the data and 83\% without. When the evidence is withdrawn or reversed, agents often notice the change but keep the original answer, and the benchmark still awards 77\% of its credit after withdrawal. Their evidence-grounded accuracy is only 41\%, and it orders the agents very differently from accuracy. Hiding gene names raises flip-following from 58\% to 93\%, supporting competition from prior knowledge as an explanation. The benchmark score and LLM judges share this blind spot.

To summarize, our contributions are as follows.
\begin{itemize}
[nosep,leftmargin=20pt,topsep=2pt,itemsep=1pt,parsep=0pt]
\item \textbf{Evaluation target.} We formalise evidence grounding through invariance and responsiveness, introduce \emph{evidence-grounded accuracy} to measure correctness together with the response to withdrawal and flip, and separate agent grounding from evaluator validity.
\item \textbf{Verified counterfactual protocol.} We build paired versions of raw data files that preserve, withdraw or reverse the evidence, verify every edit with a pre-registered statistic and control leaks through file metadata that could reveal an edit.
\item \textbf{Empirical evidence.} On single-cell tasks, we show that high accuracy can coexist with weak responsiveness to evidence and a different ranking of agents, and use gene-name anonymisation to show that prior knowledge can compete with the data.
\item \textbf{Evaluator blind spots and targeted mitigations.} The official score and LLM judges can reward answers that ignore changed evidence. An abstention option, a grounding-focused judge and transform consistency each mitigate a distinct failure, but none suffices alone.
\end{itemize}

\vspace{-5pt}
\section{Related work}
\label{sec:related}
\vspace{-5pt}

\textbf{Benchmarks for AI scientists and scientific discovery.}
Science is judged not only by its conclusions but also by how they were reached. AI scientists now reach conclusions end to end~\citep{lu2024ai,gottweis2026accelerating,yang2025r}, and as these systems proliferate, their evaluation shapes what capabilities are developed. Benchmarks have expanded from scientific knowledge~\citep{sun2024scieval,hendrycks2020measuring} and question answering~\citep{phan2025humanity,rein2023gpqa} to data-driven analysis~\citep{majumder2025discoverybench,chen2025scienceagentbench}, code execution~\citep{tian2024scicode,chan2025mle}, and research replication~\citep{siegel2024core,starace2025paperbench}.
In computational biology, recent benchmarks pair real or synthetic data with verifiable answers, covering analysis decisions~\citep{gu2024blade}, bioinformatics and study-scale analyses~\citep{mitchener2025bixbench,koch2026bixbench3}, single-cell annotation and discovery~\citep{luo2026benchmarking,workman2026scbench,diks2026scbench}, and synthetic multi-stage genetics~\citep{Li2026.06.29.735386}. These designs bring evaluation closer to how research is done. At the same time, stronger models have more ways to reach the reference answer without the data, by recall, by eliminating options~\citep{ren2026right} or by exploiting the grader~\citep{baker2025monitoring,zhong2026impossiblebench}. This raises a natural question: does a high score show that the agent's answer was determined by the data supplied in that run?

\textbf{Validity of benchmark scores.}
Whether a benchmark measures what it was built to measure is a question of benchmark validity. Prior work finds recurring problems in how LLM benchmarks define target abilities, choose tasks and score answers~\citep{bean2026measuring}, and leaderboard rankings can shift under small choices such as answer-option order or answer extraction~\citep{alzahrani2024benchmarks}.
Process evaluation grades intermediate steps rather than final answers alone, through step-level supervision~\citep{ICLR2024_aca97732}, rubrics~\citep{starace2025paperbench}, agent judges~\citep{pmlr-v267-zhuge25a} and LLM judges of analysis decisions~\citep{gu2024blade,koch2026bixbench3}.
Yet a judge sees only one run and relies on the agent's account, which may not be reliable.
LLM judges can be misled by agents' claims of task completion~\citep{lu2025agentrewardbench}, and written reasoning need not reflect what produced an answer~\citep{turpin2023language,chen2025reasoning}.
An agent can thus describe a valid analysis of the data while its final answer is not grounded in them.

\textbf{Counterfactual tests.}
A natural way to test whether an answer depends on the data is to change the evidence and see whether the answer changes.
Early work in NLP and vision showed that models can often answer without key inputs, such as the premise~\citep{poliak2018hypothesis,gururangan2018annotation}, the passage~\citep{kaushik2018much}, the question~\citep{balepur2024artifacts} or the image~\citep{chen2024we}.
Later work paired inputs with edited versions, testing whether predictions remain stable under irrelevant edits and change when relevant evidence is altered, in NLP~\citep{ribeiro2020beyond,gardner2020evaluating}, software testing~\citep{10.1145/3143561} and medical AI~\citep{gu2026evaluating,zafar2026beyond}.
Studies of knowledge conflicts find that models can hold on to what they learned in pretraining when the context contradicts it~\citep{longpre2021entity,xie2024adaptive}, and ImpossibleBench applies a similar idea to coding agents~\citep{zhong2026impossiblebench}. Scientific benchmarks have begun to adopt such tests.
DiscoveryBench compares performance with and without data~\citep{majumder2025discoverybench}, BAISBench compares against shuffled data~\citep{luo2026benchmarking}, and scBench-Long swaps labels in one test~\citep{diks2026scbench}.
These tests show that performance depends on the data on average, but not whether a particular credited answer follows the evidence, nor whether the benchmark score would still reward an agent that ignores a change.
We differ in editing the raw data files of each item, verifying every edit with a reference statistic so that the correct response is known, and testing the benchmark's score as well as the agent.

\vspace{-5pt}
\section{Preliminaries: a taxonomy of current evaluation}
\vspace{-5pt}
\label{sec:prelim}
\textbf{Setting.} A data-driven benchmark item consists of a task $x$, which may be a question or an analysis instruction, data $E$ and a reference answer, or key, $c^\star$. An agent works on the task in a loop. At each step it writes code or calls a tool, $u_i$, which is executed against the data and returns an observation $o_i$. After $n$ steps it returns a final answer,
\begin{equation}
\tau=(u_1,o_1,\dots,u_n,o_n),\qquad o_i=\mathrm{Exec}(u_i,E),\qquad a=\pi_\theta(x,\tau),
\label{eq:loop}
\end{equation}
where $\pi_\theta$ is the agent with model parameters $\theta$, and an evaluator $M$ assigns a score. The data reach the answer only through the observations, while $\theta$ carries what the model learned in pretraining, so a correct answer alone does not reveal how much the supplied data contributed.

\textbf{Current practice.} Existing evaluations fall into four families that differ in what the evaluator compares and at what level (Table~\ref{tab:taxonomy}). \textit{Outcome scoring} compares the answer with the key, $M(a,c^\star)$, and says whether one run was right. \textit{Process scoring} also reads the trajectory, $M(\tau,a,c^\star)$, but still judges one run against the same key. \textit{Input ablation} compares accuracy with and without the data, but only on average over items. \textit{Ad hoc perturbation} edits one item's data and checks whether the answer changes, but assumes the correct response to the edit and leaves the score unexamined. Single-run scores and aggregate ablations cannot establish whether a particular correct answer tracks the supplied data, and an unverified perturbation cannot establish whether a changed answer is appropriate. None of these procedures tests whether the evaluator rewards responses to verified changes in evidence.

\begin{table}[t]
\centering\small
\caption{Families of evaluation for data-driven AI scientists. Only a verified counterfactual tests single answers against a known correct response and also tests the evaluator.}
\label{tab:taxonomy}
\setlength{\tabcolsep}{4pt}
\begin{tabular}{lcccc}
\toprule
family & compares & level & correct response & evaluator \\
 & & & to the change known & tested \\
\midrule
outcome scoring & answer vs.\ key & run & -- & no \\
process scoring & trajectory vs.\ key & run & -- & no \\
input ablation & with vs.\ without data & benchmark & -- & no \\
ad hoc perturbation & before vs.\ after an edit & item & assumed & no \\
\textbf{verified counterfactual (ours)} & paired data versions & run & \textbf{verified} & \textbf{yes} \\
\bottomrule
\end{tabular}
\vspace{-8pt}
\end{table}

\vspace{-8pt}
\section{Method: evidence grounding and verified counterfactuals}
\vspace{-8pt}
\label{sec:method}
We propose to evaluate AI scientists by whether their answers follow the evidence, and to test evaluators in the same way. Our approach rests on one observation and one design principle. The observation is that whether an answer came from the data is a counterfactual property, which no evaluation of a single run can establish (Sec.~\ref{sec:grounding}). The principle is \textbf{one set of pairs, two audits}. We pair each item with versions of its data whose correct response is verified (Sec.~\ref{sec:construct}), and use the same paired runs to audit both the agent and the evaluator (Sec.~\ref{sec:scoring}).

\vspace{-5pt}
\subsection{Problem formulation}
\vspace{-5pt}
\label{sec:grounding}
We start from what the data themselves support. We define $s^\star(E)$ as the answer supported by the data, computed by a reference statistic that the task implies, for example \textit{the cell type with the highest mean expression of the queried gene}. On an item whose key is supported by its data, $s^\star(E)=c^\star$. An intervention $T$ edits the data while leaving the task and prompt unchanged, and we classify interventions by their effect on $s^\star$:
\ding{182} a \textit{null} intervention $T_0$ preserves the supported answer, $s^\star(T_0E)=s^\star(E)$,
\ding{183} a \textit{withdrawal} $T_B$ removes sufficient support for any answer, so $s^\star(T_BE)$ is undefined, and
\ding{184} a \textit{flip} $T_C$ changes the supported answer to another option, $s^\star(T_CE)=c'\neq c^\star$.
App.~\ref{app:example} works through every definition of this section on one item.

We call an answer \textit{evidence-grounded} when it tracks $s^\star$ across these interventions: it remains unchanged under $T_0$, changes or is flagged under $T_B$, and changes to $c'$ under $T_C$. The definition concerns only how the answer responds to the evidence in each run and places no constraint on the analysis, so our object of evaluation is the relation between answer and evidence rather than the analysis path, model or scaffold. In terms of Eq.~\eqref{eq:loop}, editing $E$ changes the observations $o_i$, and grounding asks whether $a$ changes accordingly. An agent may therefore detect an intervention without being grounded in it, when some $o_i$ reflects the edit but $a$ does not follow. We term this \textit{noticing without acting}.

\textbf{Why observational evaluation cannot certify grounding.}

\emph{Proposition 1 (observational non-identifiability).}
For any observed trajectory--answer pair $(\tau,a)$ on evidence $E$, there exist two agents $\pi$ and $\tilde{\pi}$ that produce exactly $(\tau,a)$ on $E$ but respond differently to some intervention $T$ on $E$. Consequently, no evaluator of the form $M(\tau,a,c^\star)$ computed only on the original evidence can determine whether the answer is evidence-grounded.

\emph{Proof sketch.}
Construct $\pi$ and $\tilde{\pi}$ to behave identically on $E$, yielding the same actions, observations, trajectory $\tau$, and final answer $a$. Let them differ only on some intervened evidence $T(E)$. For example, under a flip $T_C$, let $\pi$ change its answer to the new data-supported option $c'=s^\star(T_CE)$, while $\tilde{\pi}$ retains its original answer. Since an evaluator computed only from the run on $E$ receives identical inputs for the two agents, it must assign them the same score, although their grounding behavior differs under intervention. \hfill$\square$

The ambiguity is not hypothetical. An agent may run a correct analysis, report that the evidence is missing and still return the key (App.~\ref{app:example}), a run that looks strong under both outcome and process scoring. Separating grounded from ungrounded answers therefore requires an intervention on $E$ whose effect on $s^\star$ is known, which is what a verified counterfactual provides.

\vspace{-5pt}
\subsection{Constructing verified counterfactuals}
\vspace{-5pt}
\label{sec:construct}
\begin{figure}[t]
\centering
\includegraphics[width=\linewidth]{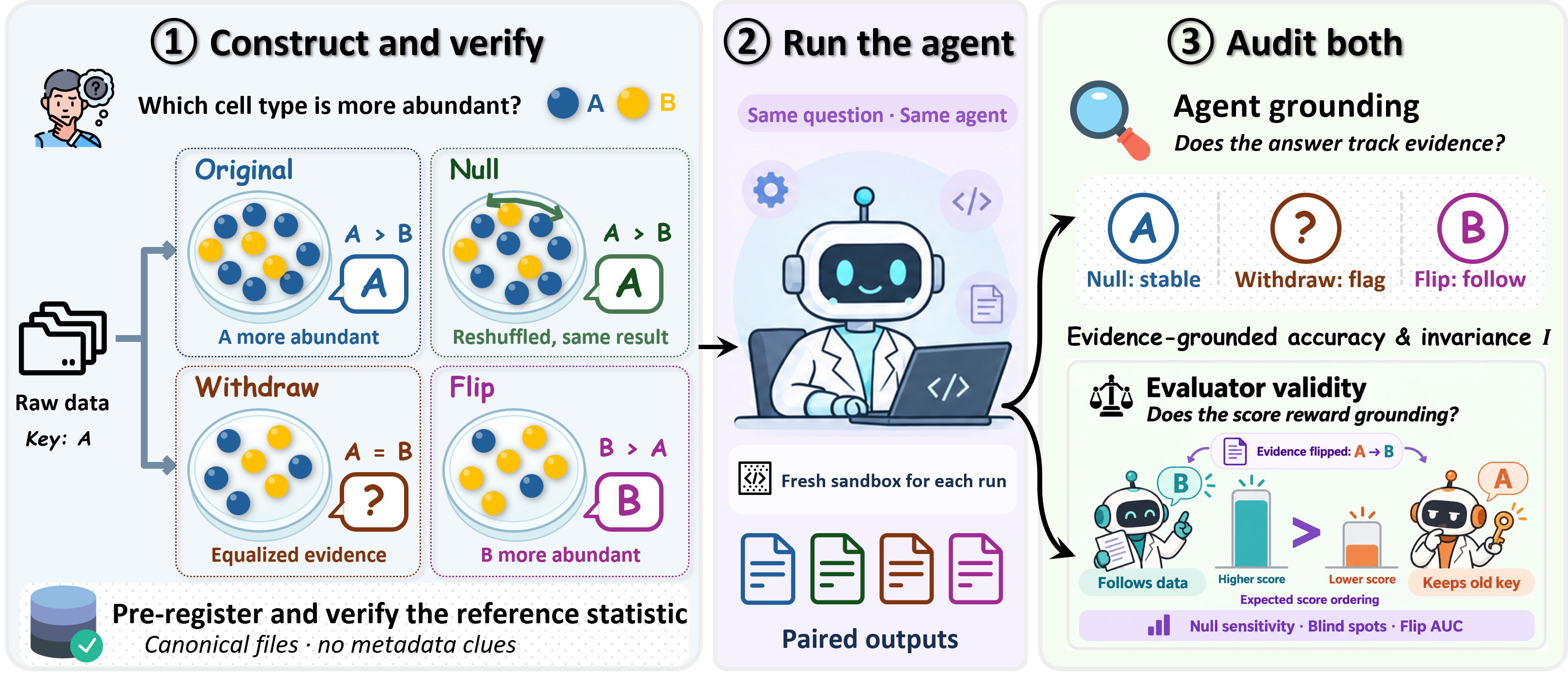}
\caption{\textbf{Protocol.} (1)~Each item is paired with null, withdrawal and flip versions of its data, each verified with a pre-registered reference statistic. (2)~The same agent runs on every version in a fresh sandbox. (3)~The paired outputs audit both the agent, which should track the evidence, and the evaluator, which should score an evidence-following answer above one that keeps the old key.}
\vspace{-10pt}
\label{fig:pipeline}
\end{figure}

For each item we pre-register the reference statistic and build one version of the data for each intervention class (Fig.~\ref{fig:pipeline}).
\textit{Null versions} permute cells and genes and rename barcodes and donors.
\textit{Withdrawal versions} delete the queried gene, remove the contrast group or equalise the candidate groups.
\textit{Flip versions} move a gene's expression to another cell type, swap two genes or down-sample one population in one condition.
Each edit touches only the quantity the statistic reads and is applied to the expression matrix, all layers and the raw counts.
A version counts as verified only when the statistic, recomputed on every file and on the raw counts, behaves as its class requires. The correct response of a grounded agent is therefore known for every version.

\textbf{Keeping the counterfactuals clean.} A counterfactual is informative only if the versions differ in evidence alone. Agents read metadata. In development one agent recovered a swap from a file name stored in a gzip header, and the h5ad files ship precomputed summaries that would still show the original evidence. Every file of every version, the original included, therefore passes through one canonical writer (App.~\ref{app:protocol}).

\vspace{-5pt}
\subsection{Scoring agents and evaluators}
\vspace{-5pt}
\label{sec:scoring}
\textbf{Agents.} Writing $a(E)$ for the answer of a run on data $E$ and $\mathrm{flag}(E)$ for an explicit statement that the data do not support an answer, we define invariance and responsiveness over paired runs,
\begin{equation}
\begin{gathered}
I=\Pr\big[a(T_0E)=a(E)\big],\qquad
R_{\text{flip}}=\Pr\big[a(T_CE)=c'\big],\\
R_{\text{rm}}=\Pr\big[a(T_BE)\neq a(E)\ \vee\ \mathrm{flag}(T_BE)\big],\qquad
R^{\text{flag}}_{\text{rm}}=\Pr\big[\mathrm{flag}(T_BE)\big].
\end{gathered}
\label{eq:metrics}
\end{equation}
We compare $I$ with the agreement between two runs on the unmodified data, which separates sensitivity to the intervention from sampling variation. $R_{\text{rm}}$ measures whether the answer depends on the withdrawn evidence at all, the counterfactual dependence of Proposition~1, while $R^{\text{flag}}_{\text{rm}}$ measures whether the response is appropriate, and we report both.

We introduce \emph{evidence-grounded accuracy} (EGA), which credits a correct answer only when it also reacts to withdrawal and follows the flip,
\begin{equation}
\mathrm{EGA}=\Pr\big[\,a(E)=c^\star\ \wedge\ \big(a(T_BE)\neq a(E) \vee \mathrm{flag}(T_BE)\big)\ \wedge\ a(T_CE)=c'\,\big].
\label{eq:ega}
\end{equation}
The strict variant $\mathrm{EGA}^{\text{strict}}$ replaces the withdrawal condition by $\mathrm{flag}(T_BE)$. We assess null invariance separately with $I$. EGA combines correctness with responsiveness to the two interventions that change the evidence. EGA does not penalise prior knowledge. It asks only whether the answer remains responsive to the evidence supplied in the current run.

Paired runs also make the route to each correct answer observable. Classifying each correct run by whether its withdrawal run \emph{reacts}, by changing the answer or flagging that the data cannot decide, and whether its flip run follows $s^\star$ to $c'$, we decompose accuracy as
\begin{equation}
\mathrm{Acc}=\underbrace{P_{\text{derived}}}_{\text{reacts, follows}}+\underbrace{P_{\text{elim}}}_{\text{persists, follows}}+\underbrace{P_{\text{rm-only}}}_{\text{reacts, keeps }c^\star}+\underbrace{P_{\text{recalled}}}_{\text{persists, keeps }c^\star}+P_{\text{other}},
\label{eq:routes}
\end{equation}
where $P_{\text{other}}$ collects flip runs that move to a third option or give no answer. Only the derived route is evidence-grounded, so $P_{\text{derived}}=\mathrm{EGA}$ and the other components are the part of the score that the data do not certify. An answer \emph{selected by elimination} persists under withdrawal because a withdrawal removes the evidence for the key but not the evidence against the other options, so in a closed option set the key can still be reached by ruling the others out. A \emph{recalled} answer persists under withdrawal and keeps $c^\star$ after a flip, consistent with knowledge outside the supplied data, and $P_{\text{rm-only}}$ reacts when the evidence disappears but not when it contradicts the key. We also report the \emph{evidence gain}, accuracy with data minus accuracy without, and, for synthetic problems whose ground truth transforms exactly under a flip, \emph{transform consistency}, which compares the flipped answer with the same transform of the agent's own original answer and so measures responsiveness even when the original answer misses the tolerance.

\textbf{Evaluators.} Throughout, the evaluator is given the original key $c^\star$, as a benchmark would be, since it is not told that the data were edited. We call an evaluator \emph{counterfactually valid} if, over paired runs, it is stable under null interventions up to run-to-run variation, does not credit $c^\star$ once the edited data no longer support it, and scores a run that follows a verified flip above one that does not. Writing $M(\cdot)$ for its score of a run against the original key, we measure the three conditions by
\begin{equation}
\begin{gathered}
S_{\text{null}}=\mathbb{E}\big|M(E)-M(T_0E)\big|,\qquad
B_{c}=\mathbb{E}\big[M(T_cE)\big]\ \ (c\in\{\text{rm},\text{flip}\}),\\
\mathrm{AUC}_{\text{flip}}=\Pr\big[M(r^+)>M(r^-)\big]+\tfrac{1}{2}\Pr\big[M(r^+)=M(r^-)\big].
\end{gathered}
\label{eq:evaluator}
\end{equation}
The null sensitivity $S_{\text{null}}$ should not exceed the change between two runs on the original data, the blind spot $B_c$ should be near zero, and the flip AUC, which compares a flip run $r^+$ answering $c'$ with one $r^-$ that does not, should be near 1. Evidence grounding is thus a property of the agent and counterfactual validity a property of the evaluator, and both are measured on the same pairs.

\vspace{-8pt}
\section{Experimental setup}
\label{sec:setup}
\vspace{-8pt}

\textbf{Tasks.} We use BAISBench's scientific-discovery split (BAIS-SD)~\citep{luo2026benchmarking} and GeneBench-Pro~\citep{Li2026.06.29.735386}. Of 193 BAIS-SD questions, 22 can be decided from the supplied single-cell data by a reference statistic. Four of these were excluded because the original data do not support the benchmark key, leaving 18 questions from 13 studies. The official score is an exact set match with half credit for a strict subset of the key. GeneBench-Pro contributes four synthetic statistical-genetics problems whose ground truth transforms exactly under specified data edits, and its grader requires every reported quantity to fall within tolerance. More details on the tasks and their interventions are provided in App.~\ref{app:tasks}.

\textbf{Agents.} We evaluate three scaffolds and five models. Claude Code runs Claude Opus 5.5, Sonnet 5 and Haiku 4.5, and Codex CLI runs GPT-5.6-Sol and GPT-5.6-Terra, each with three replicates.
Biomni~\citep{Huang2025.05.30.656746} runs Claude Sonnet 5 with one replicate. Each run starts in a fresh sandbox with only its task files and no web or retrieval access, receives the same prompt on every data version, and writes its answer, a confidence and whether the data support the answer. A no-data condition provides only the question and background. More details on the agent protocol are provided in App.~\ref{app:protocol}.

\textbf{Evaluators.} We stress-test three kinds of evaluator on the same paired runs: the benchmark's official score against the original key, an LLM process judge that reads the question, the key and a condensed trajectory (Claude Opus 5.5 and GPT-5.6-Sol), and the agent's own report of its confidence and of whether the data support its answer. Separately, an audit of all 2{,}944 included runs (16{,}091 tool calls) for access outside the sandbox or to benchmark answer material flagged none. More details on the judges and the audit are provided in App.~\ref{app:evaluators} and App.~\ref{app:protocol}.

\textbf{Research questions.} We organise the results around four questions. \emph{RQ1} asks whether agents track the evidence when it is withdrawn or reversed (Sec.~\ref{sec:rq1}). \emph{RQ2} asks how much of their accuracy is evidence-grounded (Sec.~\ref{sec:rq2}). \emph{RQ3} asks why agents fail to follow the evidence, and separates prior knowledge from analytical skill (Sec.~\ref{sec:rq3}). \emph{RQ4} asks whether the benchmark score and LLM judges notice these failures, and whether they can be repaired (Sec.~\ref{sec:rq4}).

\section{Results and analysis}
\label{sec:results}
\label{sec:evaluators}

\vspace{-4pt}
\subsection{RQ1: Do agents track the evidence?}
\label{sec:rq1}
If an agent's answers come from the data, they should stay the same under edits that leave the evidence intact and change when the evidence behind the key is withdrawn or reversed. We ran every agent on all three versions of each item under the same prompt and compare the paired answers.

\vspace{-4pt}
\begin{table}[ht]
\centering
\caption{Accuracy, invariance and responsiveness (\%). Claude rows: 18 (BAIS) or 4 (GB-Pro) tasks $\times$ 3 replicates. Accuracy: official score on $E$ and with no data. $I_{\text{rep}}$: agreement of two runs on $E$. $I_{\text{null}}$: same answer after a null edit. $R^{\text{flag}}_{\text{rm}}$: the agent reports that the withdrawn data do not support an answer. $R_{\text{flip}}$: the answer follows the flip (GB-Pro: passes against the transformed ground truth). $B$: credit against the original key after withdrawal or flip. EGA: Eq.~\eqref{eq:ega}, not defined for GB-Pro. $^\dagger$Codex CLI, 3 replicates each. $^\ddagger$Biomni on Sonnet 5, 1 replicate, no separate no-data run.}
\label{tab:main}
\resizebox{\linewidth}{!}{%
\begin{tabular}{llccccccccc}
\toprule
 & & \multicolumn{2}{c}{Benchmark accuracy} & \multicolumn{2}{c}{Invariance} & \multicolumn{2}{c}{Responsiveness} & \multicolumn{2}{c}{Blind spot $B$} & \\
\cmidrule(lr){3-4}\cmidrule(lr){5-6}\cmidrule(lr){7-8}\cmidrule(lr){9-10}
Bench. & Model & $E$ & no data & $I_{\text{rep}}$ & $I_{\text{null}}$ & $R_{\text{rm}}^{\text{flag}}$ & $R_{\text{flip}}$ & rm & flip & \textbf{EGA} \\
\midrule
\multirow{6}{*}{BAIS} & \mname{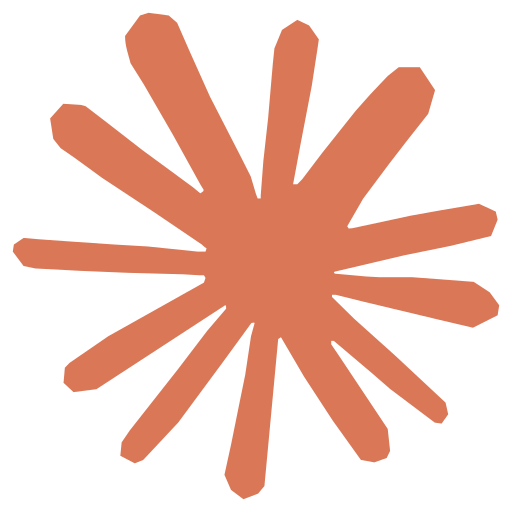}{Opus 5.5} & 98 & 85 & 96 & 98 & 41 & 59 & 85 & 33 & \textbf{35} \\
 & \mname{icon/anthropic.png}{Sonnet 5} & 100 & 89 & 100 & 100 & 31 & 52 & 89 & 41 & \textbf{31} \\
 & \mname{icon/anthropic.png}{Haiku 4.5} & 87 & 74 & 96 & 94 & 31 & 72 & 56 & 15 & \textbf{56} \\
 & \mname{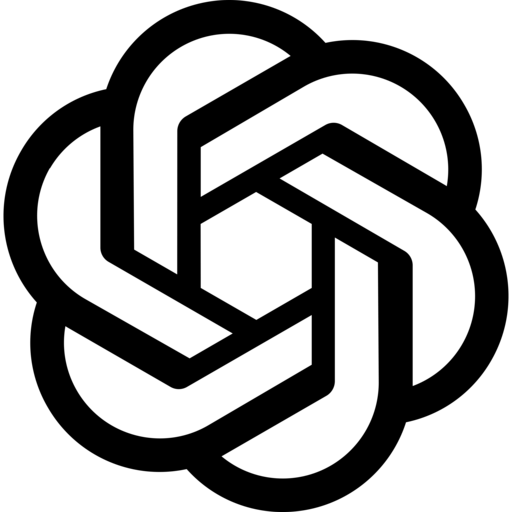}{GPT-5.6-Sol$^\dagger$} & 100 & 81 & 100 & 98 & 31 & 65 & 81 & 28 & \textbf{33} \\
 & \mname{icon/oai.png}{GPT-5.6-Terra$^\dagger$} & 94 & 80 & 93 & 94 & 37 & 63 & 78 & 31 & \textbf{35} \\
 & \mname{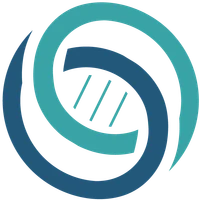}{Biomni (Sonnet 5)$^\ddagger$} & 100 & -- & 100 & 94 & 28 & 67 & 72 & 28 & \textbf{39} \\
\midrule
\multirow{3}{*}{GB-Pro} & \mname{icon/anthropic.png}{Opus 5.5} & 42 & -- & 100 & 58 & 50 & 25 & 0 & 0 & -- \\
 & \mname{icon/anthropic.png}{Sonnet 5} & 17 & -- & 25 & 8 & 25 & 8 & 0 & 0 & -- \\
 & \mname{icon/anthropic.png}{Haiku 4.5} & 0 & -- & 17 & 17 & 33 & 0 & 0 & 0 & -- \\
\bottomrule
\end{tabular}%
}
\vspace{-8pt}
\end{table}

\vspace{-5pt}\paragraph{\ding{224} Answers are stable when the evidence is intact.}
Null edits, which shuffle cells and genes and rename identifiers, change answers no more often than rerunning does (98\% agreement in both cases, Table~\ref{tab:main}), so the answer changes reported below reflect the evidence rather than chance.

\begin{figure}[ht]
\centering
\includegraphics[width=\linewidth]{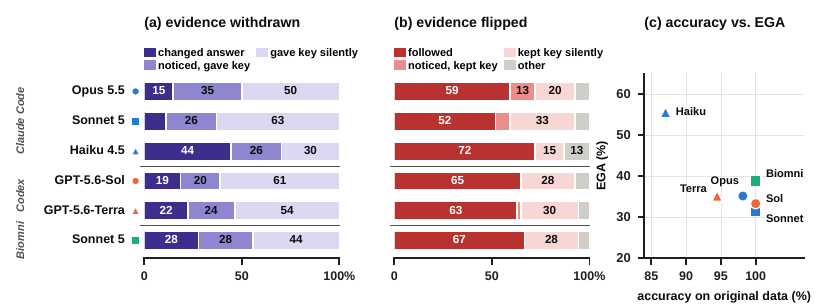}
\vspace{-16pt}
\caption{\textbf{Agents notice but do not act.} What each agent does when the key's evidence is (a)~withdrawn or (b)~flipped (\% of paired runs, complete replicates only). ``Noticed'': the agent reports that the data do not support an answer, yet gives the key. (c)~Accuracy on the original data against EGA. The two orderings differ markedly. Colour: scaffold. Marker: model (Biomni runs Sonnet 5).}
\label{fig:behavior}
\vspace{-8pt}
\end{figure}

\vspace{-5pt}\paragraph{\ding{224} Withdrawn evidence is noticed but not acted on.}
When the evidence behind the key is withdrawn, the benchmark still awards 77\% of its credit to the Claude agents (Table~\ref{tab:main}). The agents do register the loss (App.~\ref{app:analyses}), yet only 35\% of withdrawal runs state that the data cannot decide, and 84\% of those give the key regardless. As Fig.~\ref{fig:behavior}a shows, the forced choice turns this uncertainty into a committed answer, which the trajectories show is often reached by ruling out the other options or from memory (App.~\ref{app:example}, App.~\ref{app:cases}). The benchmark score cannot tell such an answer from analysis.

\vspace{-5pt}\paragraph{\ding{224} Even reversed evidence often fails to move the answer.}
A reversal is a sharper test, because the data now point to a different option. Only 61\% of runs follow it, and in 30\% the agent keeps the key and is still credited (Fig.~\ref{fig:behavior}b). A scientist might reasonably distrust data that contradict known biology, but most runs that keep the key never mention a conflict, and reversals that look biologically ordinary, such as a shift in cell-type proportions, are followed only slightly more often (67\%). The effect holds in every scaffold and without the two tasks that admit a second valid analysis (App.~\ref{app:tasks}, App.~\ref{app:analyses}). The original answer remains sticky even when the supplied evidence supports another option.

\vspace{-4pt}
\subsection{RQ2: How much of the accuracy is evidence-grounded?}
\label{sec:rq2}
Accuracy credits every correct answer alike. Using the paired runs, we ask how much of it remains once a correct answer must also respond to both edits, and by which routes the rest is reached.

\vspace{-4pt}
\begin{figure}[ht]
\centering
\includegraphics[width=\linewidth]{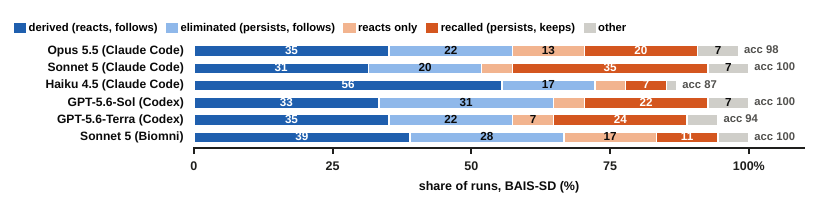}
\vspace{-10pt}
\caption{\textbf{Where the accuracy comes from.} Each correct BAIS-SD run is classified by its paired withdrawal and flip runs (Eq.~\ref{eq:routes}). Only the dark-blue segment, runs that react to withdrawal and follow the flip, is evidence-grounded accuracy. Bars sum to accuracy. Complete replicates only.}
\label{fig:routes}
\vspace{-8pt}
\end{figure}

\vspace{-5pt}\paragraph{\ding{224} Most correct answers do not meet both response conditions.}
By this standard the Claude agents fall from 95\% accuracy to 41\% EGA, and every agent loses more than a third of its accuracy (Table~\ref{tab:main}). Figure~\ref{fig:routes} breaks down the rest. Fewer than half of the correct runs react to withdrawal and follow the flip (the derived route), roughly one in five follows the elimination pattern and a similar share the recall pattern. The two measures also order the agents differently, as Fig.~\ref{fig:behavior}c shows. Sonnet and GPT-5.6-Sol answer every item correctly yet have the lowest EGA, while Haiku, the least accurate, has the highest (Kendall's $\tau=-0.54$, not significant with six agents). Part of Haiku's lead comes from changing its answer after withdrawal without flagging missing evidence. The strict variant, which requires such a flag, reduces all agents to 17--28\% and leaves the ordering inverted.

\begin{figure}[t]
\centering
\includegraphics[width=\linewidth]{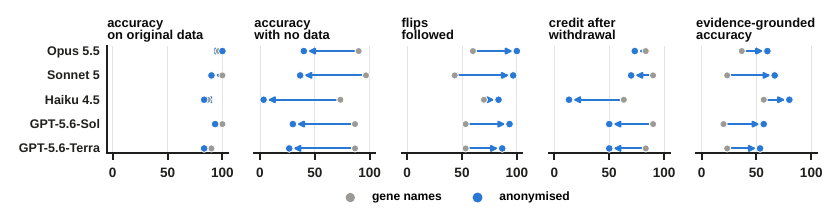}
\vspace{-12pt}
\caption{\textbf{Gene-name anonymisation.} The same 10 gene tasks with every gene symbol replaced by an opaque identifier in data and text (grey: names, blue: anonymised). The no-data prior falls towards chance, accuracy with data barely changes, and the same agents now follow the flipped evidence.}
\label{fig:anon}
\vspace{-12pt}
\end{figure}

\vspace{-4pt}
\subsection{RQ3: Why do agents not follow the evidence?}
\label{sec:rq3}
An agent may fail to follow the evidence because it cannot analyse the data well, or because what it already knows overrides the data. We separate them by removing the prior but not the evidence.

\vspace{-5pt}\paragraph{\ding{224} No-data accuracy and the recall pattern covary.}
Given only the question and background, the Claude models answer 83\% correctly and the GPT models about as many, so the data add only 12 percentage points. Sonnet answers the most questions without data and shows the recall pattern most often, and Haiku does both least (Fig.~\ref{fig:routes}), suggesting that the prior competes with the data.

\vspace{-5pt}\paragraph{\ding{224} Hiding gene names makes agents follow reversed evidence.}
We replaced every gene symbol with an opaque identifier in the 10 gene tasks, consistently in the data, question, options and background (App.~\ref{app:anon}). This preserves the underlying evidence while removing gene identity as a usable prior. As Fig.~\ref{fig:anon} shows, accuracy without data falls from 87\% to 27\%, near chance, while accuracy with data barely changes, so the tasks remain solvable from the data alone. Under reversal, agents now follow the flipped evidence in 93\% of runs instead of 58\%, and EGA rises from 39\% to 69\%. The GPT agents change the same way, and the effect remains without the one task whose study the models still recognise. Anonymisation also removes semantic cues, so this supports prior competition as an important explanation rather than proving it the only one. Withdrawal improves less (key credit 79\% to 52\%), consistent with the forced choice still allowing answers by elimination.

\vspace{-5pt}\paragraph{\ding{224} Without a competing prior, analysis quality becomes the limit.}
GeneBench-Pro offers little task-specific prior to fall back on. There, the agents' answers on flipped data transform with the ground truth about as often as repeated runs on unchanged data agree, although accuracy stays low (App.~\ref{app:analyses}). Without a competing prior, responsiveness appears limited mainly by analytical precision.

\vspace{-4pt}
\subsection{RQ4: Do evaluators notice, and can they be repaired?}
\label{sec:rq4}
\label{sec:repairs}
If agents often ignore changed evidence, a valid evaluator should notice. We applied the stress test of Sec.~\ref{sec:scoring} to the benchmark score, LLM judges and the agents' self-reports, and tested targeted repairs.

\begin{figure}[t]
\centering
\includegraphics[width=\linewidth]{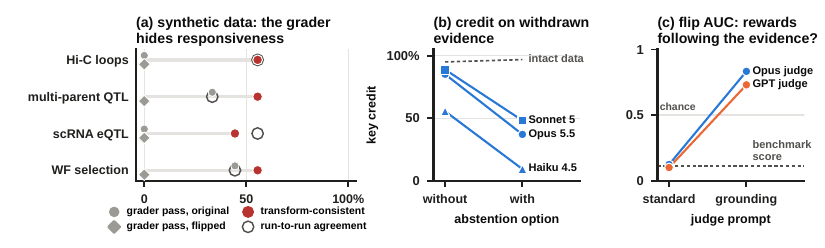}
\vspace{-10pt}
\caption{\textbf{What the score hides, and what repairs it.} (a)~GeneBench-Pro (Claude models). The grader passes almost no run on either version, yet transform-consistent answers are as frequent as run-to-run agreement. (b)~Offering ``the provided data are insufficient'' cuts key credit on withdrawn evidence for Claude Code agents and leaves intact-data accuracy unchanged. (c)~Flip AUC, the probability that a run following the flip outscores one that does not. Under the standard prompt both judges reward ignoring the flip, like the benchmark score. The grounding prompt reverses this.}
\label{fig:repairs}
\vspace{-14pt}
\end{figure}

\vspace{-5pt}\paragraph{\ding{224} Neither the score nor LLM judges reward following the evidence.}
The benchmark score is stable under null edits, but falls under flips whenever the answer leaves the key, so a run that follows the flipped data scores below one that ignores them nearly nine times in ten (flip AUC 0.11). An LLM judge sees the condensed analysis, including statements that evidence is missing, yet grades against the same key and behaves alike (flip AUC 0.12). A GPT judge does the same (0.10), so the blind spot is not family-specific (App.~\ref{app:evaluators}). Self-reports separate the two cases better, since runs that keep the key after a flip rarely claim full support (13\%), unlike runs that follow it (84\%). Synthetic data reverse the failure. As Fig.~\ref{fig:repairs}a shows, the grader scores most runs zero on every version, so a run tracking the transformation but missing the tolerance looks like one that ignores the edit.

\vspace{-5pt}\paragraph{\ding{224} Targeted repairs fix one failure at a time.}
An explicit ``data are insufficient'' option leaves intact-data accuracy unchanged and more than halves withdrawal credit, but fails under flips, where the data support an answer (Fig.~\ref{fig:repairs}b). Asking judges to grade grounding rather than agreement lifts their flip AUC above 0.7 at no cost on null edits, but leaves part of the withdrawal blind spot (Fig.~\ref{fig:repairs}c). Transform consistency credits synthetic answers that move with the data despite missing the tolerance. No repair covers every failure, and none can be validated without paired counterfactuals (App.~\ref{app:repairs}).

\vspace{-8pt}

\section{Conclusion}
\label{sec:conclusion}
\vspace{-8pt}
Science advances when evidence can change a conclusion, yet high accuracy on a data-driven benchmark does not establish that an AI scientist used the supplied data. We tested this assumption with verified, item-level counterfactuals that preserve, withdraw or reverse the evidence. On our tasks, agents often retained the original answer after the evidence changed, while benchmark scores and LLM judges continued to reward it: Claude agents averaged 95\% accuracy but only 41\% evidence-grounded accuracy. Hiding gene names increased responsiveness to reversed evidence, suggesting that prior knowledge can override evidence the agents are able to analyse. As models learn more, accuracy may increasingly measure what an agent knows rather than what it can infer. Evaluations of data-driven AI scientists should therefore test whether both answers and scores respond to verified changes in evidence, and report evidence-grounded accuracy alongside conventional accuracy.

\newpage
\bibliographystyle{IEEEtran}
\bibliography{refs.bib}


\newpage
\appendix

\section{Discussion}
\label{app:discussion}
\textbf{What the score measures.}
Benchmark accuracy measures whether an answer meets the scoring criteria, but does not establish whether it depends on the supplied data. EGA measures behaviour under our interventions rather than the reasoning inside a run, so the routes in Eq.~\eqref{eq:routes} are behavioural patterns consistent with recall or elimination, not observations of them. For withdrawal, the plain variant accepts either a changed answer or an explicit statement that the data cannot decide, and the strict variant requires the latter.

\textbf{What the evaluator results show.} All runs were scored against the original key, as in normal use. The exact-match score still uses the original key after a flip, so it penalises answers that follow the changed data. It therefore measures agreement with the key, not whether the answer follows the supplied evidence. LLM judges could in principle credit an analysis that contradicts the key, yet under the standard prompt they followed it, and only a grounding prompt changed this (App.~\ref{app:repairs}).

\textbf{Recommendations.} We recommend four practices. Data-driven benchmarks should ship each item with a verified withdrawal and flip and report both EGA variants and the no-data prior next to accuracy. They should offer an explicit ``insufficient data'' option. They should validate process judges on counterfactual pairs and credit transform consistency where the ground truth transforms exactly. Last, they should pass every file version, the original included, through one canonical writer.

\textbf{Limitations.} Our study was limited mainly by compute and budget. Each counterfactual is built and verified by hand and every version requires a full agent run, so we covered 18 BAIS-SD questions and four GeneBench-Pro problems, ran three replicates per model, a single replicate for Biomni, and tested gene-name anonymisation on ten tasks only. Larger task sets, more replicates, open-weight models and a human validation of the LLM judges would tighten our estimates. Finally, responsiveness to evidence is necessary for data-driven reasoning but not sufficient for scientific quality.

\section{A worked example of the definitions}
\label{app:example}
This section follows one BAIS-SD item through every definition of Sec.~\ref{sec:method}. It is the item described in the introduction, and all answers below are from our runs.

\textbf{The item.} The task $x$ asks ``Which of the following gene expression markers was used to identify endothelial cells in the human hamstring tendon?'' with options A)~PDGFRB, B)~TRDN, C)~PECAM1 and D)~PAX7. The data $E$ are the single-nucleus RNA-seq file of the study~\citep{Mimpen2022.12.19.521110} and the key is $c^\star=\text{C}$.

\textbf{The reference statistic and $s^\star$.} The task asks for a marker of one cell type, so the statistic is, for each option gene, its mean log-normalised expression in endothelial cells minus its mean in all other cells. The data-supported answer $s^\star(E)$ is the option with the largest value. On the original file this is PECAM1, so $s^\star(E)=c^\star=\text{C}$ and the key is supported by the data.

\textbf{The three interventions.} The null version $T_0E$ permutes the order of cells and genes and renames cell barcodes and donors. The statistic is unchanged, so $s^\star(T_0E)=\text{C}$. The withdrawal $T_BE$ deletes PECAM1 from the expression matrix, every layer and the raw counts. The statistic can no longer be computed for the key, so the data no longer decide the task and $s^\star(T_BE)$ is undefined. The flip $T_CE$ swaps the values of PECAM1 and PAX7 in every cell. PAX7 now has the endothelial profile, so $s^\star(T_CE)=\text{D}=c'$. We confirm these outcomes by recomputing the reference statistic on every file and on the raw counts.

\textbf{The loop and noticing without acting.} On $T_BE$, one observation $o_i$ in the trajectory of Opus~5.5 leads it to state that ``PECAM1 is not in the dataset's gene list''. Its final answer is nevertheless C, ``the only option left''. The edit reached the observations but not the answer. This is noticing without acting in the sense of Sec.~\ref{sec:grounding}. On $E$ its trajectory and answer are those of a careful analyst.

\textbf{Agent metrics and routes.} Table~\ref{tab:example} lists the answers of every run on this item. All agents answer C on $E$ and on $T_0E$, so accuracy and invariance are both 100\%. The runs then separate by route. Haiku flags the withdrawn evidence and follows the flip to D in all three replicates, which is the \emph{derived} pattern. Two GPT-5.6-Terra runs give C without a flag under withdrawal but follow the flip, which is the pattern of \emph{elimination}. The other options are well-known markers of pericytes, muscle and satellite cells, so C can be chosen without PECAM1 being measured. The remaining eleven runs keep C under both edits, which is the \emph{recall} pattern. No run on this item reacts to withdrawal alone. Accuracy on the item is therefore 16/16, $R_{\text{rm}}=3/16$, $R_{\text{flip}}=5/16$ and EGA $=3/16$. The gap between 16 and 3 is exactly the elimination and recall mass of Eq.~\eqref{eq:routes}.

\textbf{Evaluator metrics.} The benchmark scores by exact match with the key C. On the null pairs every score is unchanged, so its sensitivity is zero, as it should be. On the withdrawal versions it gives full credit to all sixteen runs, including the three Haiku runs that stated the data cannot decide, because they still selected C when forced to choose. On the flip versions it gives full credit to the eleven runs that kept C and zero to the five that followed the evidence to D. Its blind spot on this item is 100\% after withdrawal and 11/16 after the flip, and its flip AUC is 0. Every run that follows the evidence scores below every run that does not. A counterfactually valid evaluator would reverse this order.

\begin{table}[h]
\centering\small
\caption{All runs on the worked example (S23\_Q3). Answers on the original data $E$, the null version, the withdrawal (with the agent's data-support label) and the flip, and the resulting route. The key is C on $E$, and the flip moves the data-supported answer to D. Rows group runs with identical answers. The three replicates of GPT-5.6-Terra split into two rows, and Biomni was run once.}
\label{tab:example}
\setlength{\tabcolsep}{4pt}
\begin{tabular}{llccccl}
\toprule
agent & runs & $E$ & $T_0E$ & $T_BE$ & $T_CE$ & route \\
\midrule
Haiku 4.5 & 3 & C & C & C, flagged & D & derived \\
GPT-5.6-Terra & 2 & C & C & C, partial support & D & eliminated \\
GPT-5.6-Terra & 1 & C & C & C, partial support & C & recalled \\
Opus 5.5 & 3 & C & C & C, partial support & C & recalled \\
Sonnet 5 & 3 & C & C & C, partial support & C & recalled \\
GPT-5.6-Sol & 3 & C & C & C, partial support & C & recalled \\
Biomni (Sonnet 5) & 1 & C & C & C, partial support & C & recalled \\
\bottomrule
\end{tabular}
\end{table}

\section{Benchmarks and tasks}
\label{app:tasks}
\textbf{BAISBench.} BAISBench~\citep{luo2026benchmarking} evaluates AI scientists on omics data-driven biological discovery with two tasks, cell type annotation and scientific discovery. Its scientific-discovery split (BAIS-SD) poses 193 multiple-choice questions derived from conclusions reported in 41 published single-cell studies. Each question comes with the study's h5ad files and a short background, and one or more options form the key. The official score is an exact set match with half credit for a strict subset of the key. Of the 193 questions, 22 can be decided from the supplied data by a single reference statistic, and 18 remain after the exclusions below.

\textbf{GeneBench-Pro.} GeneBench-Pro~\citep{Li2026.06.29.735386} evaluates multistage statistical reasoning in genomics, quantitative biology and translational biomedicine with 129 evaluations. Each problem provides a synthetic dataset generated with known ground truth, brief context and a target quantity, and an answer passes only if every reported quantity falls within a problem-specific tolerance. We use the four of its ten public problems for which a data edit has an exactly known effect on the ground truth (Table~\ref{tab:gbint}). For the other six, no such edit was available.

The 18 BAIS-SD tasks fall into five evidence families: cell-type abundance, composition change between conditions, expression localisation, marker gene and differential expression between conditions. The four GeneBench-Pro problems cover Hi-C loop strength, multi-parent QTL mapping, single-cell eQTL and Wright--Fisher selection, and their ground truth transforms exactly under a case and control swap, a founder relabelling, an allele recoding or a locus swap. Table~\ref{tab:tasks} lists the 18 BAIS-SD tasks with their withdrawal and flip operations. All tasks share the null operations: permutation of cells and genes (applied to \texttt{X}, all layers and \texttt{.raw}), opaque barcodes and opaque donor and sample identifiers. Every version of every task is verified against the reference statistic on each staged file and on the \texttt{.raw} counts.

\textbf{Excluded candidates.} Four candidates were dropped because the original data do not support the benchmark's key. In S17\_Q1, ACE2 expression in AT2 and club cells is tied. In S09\_Q5, APOA4 is absent from the files. In S32\_Q2, OLIG1 marks OPCs more strongly than BCAN. In S36\_Q3, ISG15 is more down-regulated than IFNG.

\textbf{Alternative analysis paths.} Two tasks admit a second defensible analysis. In S36\_Q1, disease activity is encoded in several \texttt{obs} columns, and in S09\_Q1 the ``Progenitor'' cells can be read as enterocyte lineage. Without these two tasks, flips are followed in 65\% of runs and withdrawal credit is 76\%, compared with 61\% and 77\% on all tasks.

\begin{table}[h]
\centering\scriptsize
\caption{BAIS-SD tasks. ``new'' is the option supported by the flipped data.}
\label{tab:tasks}
\resizebox{\linewidth}{!}{%
\begin{tabular}{p{1.2cm}p{1.5cm}p{0.5cm}p{4.3cm}p{4.3cm}p{0.5cm}}
\toprule
qid & family & key & withdrawal ($T_B$) & flip ($T_C$) & new \\
\midrule
S09\_Q1 & abundance & B & equalise option abundances & down-sample Enterocyte & A \\
S07\_Q1 & abundance & C & equalise option abundances & down-sample mesothelial cell & D \\
S35\_Q1 & abundance & C & equalise option abundances & down-sample Pericyte, Myofibroblast, Matrix\_Fibroblast, Smooth\_Muscle\_Cell, Stromal\_Cell & A \\
S06\_Q2 & abundance & C & remove cells with cellxgene\_cell\_type = T cell, natural killer cell, B cell, innate lymphoid cell & remove cells with cellxgene\_cell\_type = natural killer cell, B cell, innate lymphoid cell & A \\
S27\_Q1 & composition change & C & remove cells with Disease group = mild COVID-19, mild COVID-19 (asymptomatic) & down-sample classical monocyte and down-sample natural killer cell & B \\
S19\_Q2 & composition change & A & remove cells with disease = normal & down-sample AT1 and down-sample Endothelial cells & B \\
S36\_Q1 & composition change & A & remove cells with activity\_python\_binary\_crp = Yes & down-sample CD4 T central memory and down-sample Naive Bcells & C \\
S03\_Q3 & expression localisation & B & delete gene TIMP3 & move TIMP3 expression macroglial cell$\leftrightarrow$retina horizontal cell & A \\
S38\_Q3 & expression localisation & D & delete gene RHO & move RHO expression retinal rod cell$\leftrightarrow$retinal cone cell & A \\
S00\_Q4 & expression localisation & D & delete gene TMPRSS2 & move TMPRSS2 expression keratinizing barrier epithelial cell$\leftrightarrow$club cell & A \\
S09\_Q2 & marker gene & A & delete gene NUSAP1 & swap NUSAP1$\leftrightarrow$SOX9 & C \\
S09\_Q4 & marker gene & A & delete gene ITLN1 & swap ITLN1$\leftrightarrow$MYC & C \\
S07\_Q2 & marker gene & B & delete gene WT1 & swap WT1$\leftrightarrow$CDH1 & D \\
S35\_Q2 & marker gene & B & delete gene HES1 & swap HES1$\leftrightarrow$PTPRC & D \\
S23\_Q3 & marker gene & C & delete gene PECAM1 & swap PECAM1$\leftrightarrow$PAX7 & D \\
S36\_Q2 & marker gene & B & delete gene ISG15 & swap ISG15$\leftrightarrow$CXCR4 & D \\
S16\_Q1 & condition DE & A & remove cells with MUTATION.GROUP = none & swap CXCL3$\leftrightarrow$IL10 & D \\
S21\_Q2 & condition DE & B & remove cells with disease = pulpitis & swap VCAM1$\leftrightarrow$COL1A1 & C \\
\bottomrule
\end{tabular}%
}
\end{table}

\begin{table}[h]
\centering\small
\caption{GeneBench-Pro problems and interventions. Null versions shuffle line order in every file. The flip changes the ground truth by a known transform.}
\label{tab:gbint}
\resizebox{\linewidth}{!}{%
\begin{tabular}{llll}
\toprule
problem & withdrawal ($T_B$) & flip ($T_C$) & effect on ground truth \\
\midrule
Hi-C loop strength & case replicates withheld & case and control swapped & $\Delta$ negated \\
multi-parent QTL & phenotypes permuted across lines within batch & founder labels permuted & high founder F5 becomes F2 \\
scRNA eQTL & donor genotypes permuted & alleles recoded, $g\to2-g$ & $\beta$ negated \\
Wright--Fisher selection & locus A's trajectory replaced by B's & loci A and B swapped & selected locus becomes B \\
\bottomrule
\end{tabular}%
}
\end{table}

\section{Agent protocol}
\label{app:protocol}
\textbf{Data.} Before any intervention, every version, the original included, passes through one canonical writer. It drops precomputed per-gene statistics, \texttt{varm} loadings and analysis results in \texttt{uns}, and keeps the matrices, \texttt{.raw}, layers, cell annotations and embeddings. Gene edits are applied to \texttt{X}, \texttt{.raw} and all layers. Tables are re-encoded with canonical gzip headers (the file's own name and a zero timestamp). Files are hard-linked into the sandbox under the benchmark's original names.

\textbf{Common to all agents.} Each run starts in a fresh sandbox with an opaque random path under a non-listable root and sees only its task files. It has no web, search or retrieval tools and no user settings, skills, memories or MCP servers. The prompt is identical across the versions of a task. Each run has a 40-minute limit. The agent writes \texttt{answer.json} with the answer, a confidence from 0 to 100, a data-support judgement (supports, partially supports, does not support, cannot determine) and a short statement of evidence.

\textbf{Claude Code} runs headless (\texttt{claude -p}) with only the Bash, Read, Write, Edit, Glob and Grep tools, a private temporary directory and background tasks disabled.

\textbf{Codex CLI} runs \texttt{codex exec} with web search off and every stateful or web-capable feature disabled explicitly (memories, browser and computer use, in-app browser, apps, skill search and goals). A check of all Codex transcripts found only shell commands, file edits and messages.

\textbf{Biomni} (A1 agent, version 0.0.8) runs with its own loop, system prompt, know-how documents, tool registry and tool retriever, the model's API defaults and its own output limit, and without the data lake, MCP servers or self-critique. We made two changes. Biomni appends code-execution results as assistant turns, which current Claude models reject, so these results are sent in the user role with unchanged content. And network access is disabled, since name resolution is refused for every host except the model API and child processes receive a dead HTTP proxy. Biomni runs in its own Python environment with every analysis library pinned to the version used by the other agents.

\textbf{Audit.} Every transcript is checked for paths outside the sandbox (other runs' workspaces, the stored data versions and our project files, which contain the answer keys), parent-directory traversal, filesystem-wide searches, benchmark answer material and network use (URLs, HTTP libraries and package installation). No run in the analysis was flagged.

\section{Evaluators}
\label{app:evaluators}
\label{app:followup}
\textbf{Process judge.} The judge mimics current process judges~\citep{koch2026bixbench3}. It receives the task, the benchmark's reference answer, the agent's condensed trajectory (tool calls with truncated outputs, in order) and the final answer, and it is not told that the data may have been edited. Each agent message, tool call and tool output is cut to its first 1{,}200 characters, and a trajectory longer than 40{,}000 characters keeps its first and last 20{,}000. It returns whether the answer matches the reference, the quality of the executed analysis (0 to 10), whether the results the agent obtained support its answer (yes, partially or no) and an overall score (0 to 10). We analyse the overall score, rescaled to $[0,1]$, and the support rating. The main judge is Claude Opus 5.5.

\textbf{Cross-family judge.} A GPT-5.6-Sol judge graded the first replicate of every Claude and GPT-5.6-Sol run under both prompts. Its agreement with the Opus judge on the overall score is 0.91 (Spearman) and 0.89 (Krippendorff's $\alpha$, $n=306$). On whether the answer is supported, weighted $\kappa$ is 0.74, and on whether it is correct, $\kappa$ is 1.00. Under the grounding prompt the values are 0.89, 0.87 and 0.85 ($n=303$). The agreement sets contain the 288 first-replicate runs (18 tasks, four agents, four data versions) and 18 later-replicate runs that both judges also graded, and three runs lack a grounding-prompt grade. Table~\ref{tab:judges} gives the stress test for each judge.

\begin{figure}[h]
\centering
\includegraphics[width=\linewidth]{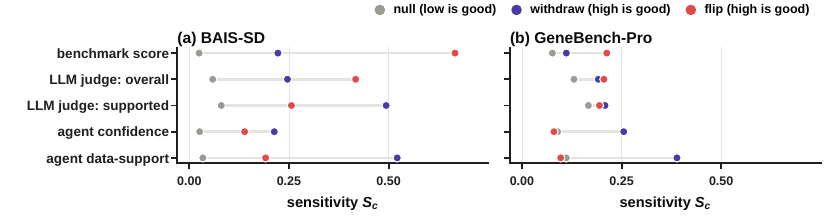}
\caption{\textbf{Evaluator stress test} (Claude Code runs). Change in each evaluator's score under null, withdrawal and flip interventions. The null change should stay near the change between replicates. How large the withdrawal and flip changes should be depends on how many runs follow each edit, so the flip AUC (Table~\ref{tab:judges}) is the direct test of whether an evaluator rewards following the evidence.}
\label{fig:eval}
\end{figure}

\textbf{Evaluator sensitivities} (Fig.~\ref{fig:eval}). These values are computed over all runs of the three Claude models with three replicates, so they differ slightly from Table~\ref{tab:judges}, which uses the first replicate of the Claude and GPT-5.6-Sol runs graded by all four judge variants. The benchmark score is as stable under null edits as between replicates (0.02 and 0.02), and changes strongly under flips (0.67) and weakly under withdrawal (0.22). The standard judge is barely more sensitive to withdrawal (0.25) and less sensitive to flips (0.42). Its sub-score for whether the answer is supported by the agent's analysis tracks withdrawal (0.49) but not flips (0.26). On GeneBench-Pro the grader changes only weakly under flips (0.21, against 0.08 under null edits), because it scores most runs zero on every version.

\begin{table}[h]
\centering\small
\caption{Evaluator stress test by judge on the same runs, the first replicate of the Claude and GPT-5.6-Sol agents graded by all four judge variants ($n$ per condition). The standard judge sees the reference answer. The grounding judge is also told that matching the reference is not sufficient. Flip AUC is the probability that a run following the flip outscores one that does not.}
\label{tab:judges}
\begin{tabular}{lccccc}
\toprule
evaluator & $S_{\text{null}}$ & $S_{\text{rm}}$ & $S_{\text{flip}}$ & flip AUC & $n$ \\
\midrule
Opus judge & 0.06 & 0.25 & 0.44 & 0.12 & 75 \\
GPT judge & 0.05 & 0.31 & 0.56 & 0.10 & 75 \\
Opus judge, grounding prompt & 0.07 & 0.43 & 0.32 & 0.83 & 75 \\
GPT judge, grounding prompt & 0.09 & 0.51 & 0.38 & 0.73 & 75 \\
benchmark score & 0.03 & 0.21 & 0.68 & 0.11 & 75 \\
\bottomrule
\end{tabular}

\end{table}

\vspace{-8pt}
\section{Additional analyses}
\label{app:analyses}
\vspace{-8pt}
Numbers in this section refer to the three Claude models with 3 replicates unless stated otherwise.

\textbf{By evidence family} (Table~\ref{tab:family}). Marker-gene questions have the highest no-data accuracy. Almost all are answerable without data, agents flag a deleted marker in only 24\% of runs, and 69\% of the withdrawal credit survives. Agents answer them by elimination among the remaining options, or say that the swapped labels look wrong and keep the canonical marker. Withdrawal is flagged most often for abundance questions (53\%). Differential-expression questions between conditions follow flips least.

\begin{table}[h]
\centering\small
\caption{Results by evidence family (\%, Claude models).}
\label{tab:family}
\begin{tabular}{lcccccccc}
\toprule
family & tasks & acc. $E$ & no data & $I_{\text{null}}$ & $R^{\text{flag}}_{\text{rm}}$ & $B_{\text{rm}}$ & $R_{\text{flip}}$ & $B_{\text{flip}}$ \\
\midrule
abundance & 4 & 100 & 81 & 100 & 53 & 64 & 67 & 33 \\
composition change & 3 & 89 & 67 & 96 & 19 & 78 & 67 & 0 \\
expression localisation & 3 & 96 & 74 & 96 & 52 & 96 & 81 & 19 \\
marker gene & 6 & 98 & 96 & 98 & 24 & 69 & 59 & 37 \\
condition DE & 2 & 83 & 83 & 94 & 28 & 94 & 17 & 61 \\
\bottomrule
\end{tabular}

\end{table}

\begin{figure}[h]
\centering
\includegraphics[width=0.6\linewidth]{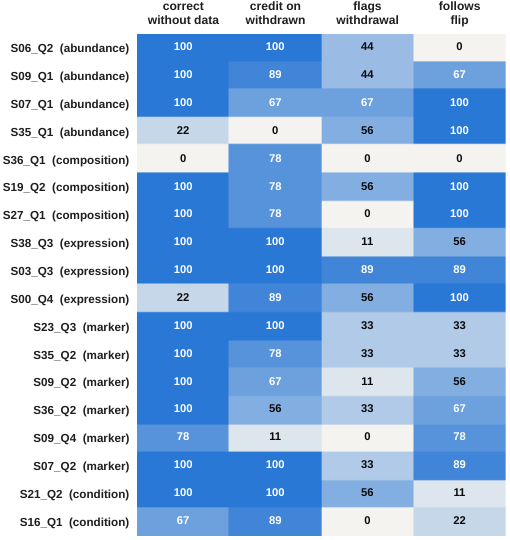}
\caption{Per-task rates (\%, mean over Claude models and replicates), grouped by evidence family.}
\label{fig:tasks}
\end{figure}

\textbf{Per task} (Fig.~\ref{fig:tasks}). The headline results are not driven by a few items. Withdrawal leaves full credit in every run for 6 of the 18 tasks, and 2 tasks never follow the flip. Of the three tasks rarely answerable without data (at most 22\%), two (S35\_Q1 and S00\_Q4) follow every flip, and the third (S36\_Q1) is a task with an alternative analysis path.

\begin{figure}[h]
\centering
\includegraphics[width=\linewidth]{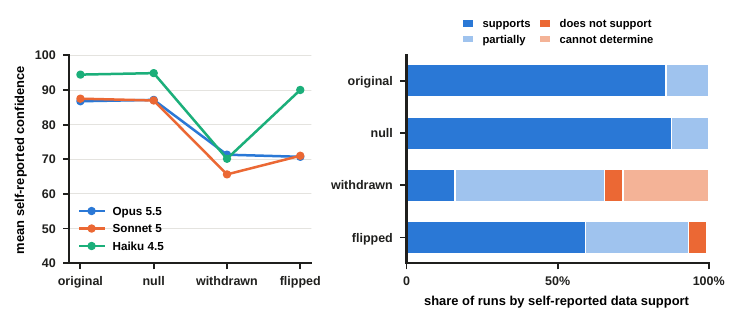}
\caption{Self-reported confidence (left) and data support (right) by data version.}
\label{fig:selfreport}
\end{figure}

\textbf{Self-reports} (Fig.~\ref{fig:selfreport}). Mean confidence falls from 90 on intact data to 69 after withdrawal, and only 16\% of withdrawal runs still report full data support. After a flip, confidence stays at 77 and 59\% of runs report full support. This share mixes two cases. Runs that follow the flip report full support in 84\% of cases, as they should, while runs that keep the key do so in only 13\%.

\textbf{Uncertainty} (Table~\ref{tab:ci}). Intervals come from a bootstrap that resamples tasks, keeping each task's models and replicates together. The blind spot on withdrawn evidence and the partial response to flips hold for every model, both GPT models included.

\begin{table}[h]
\centering\small
\caption{Per-model estimates with task-clustered 95\% bootstrap intervals (\%).}
\label{tab:ci}
\begin{tabular}{lcccc}
\toprule
model & acc. on $E$ & $B_{\text{rm}}$ & $R_{\text{flip}}$ & $B_{\text{flip}}$ \\
\midrule
Opus 5.5 & 98 [94, 100] & 85 [70, 98] & 59 [39, 80] & 33 [15, 54] \\
Sonnet 5 & 100 [100, 100] & 89 [76, 100] & 52 [30, 74] & 41 [19, 63] \\
Haiku 4.5 & 87 [70, 100] & 56 [39, 74] & 72 [52, 91] & 15 [0, 31] \\
GPT-5.6-Sol & 100 [100, 100] & 81 [63, 96] & 65 [44, 85] & 28 [11, 48] \\
GPT-5.6-Terra & 94 [85, 100] & 78 [61, 93] & 63 [41, 83] & 31 [13, 52] \\
\bottomrule
\end{tabular}

\end{table}

\textbf{Analysis effort} (Table~\ref{tab:effort}). Median wall time rises from 30\,s on intact data to 49\,s after withdrawal and 38\,s after a flip. After withdrawal the extra analysis usually confirms that the gene or group is missing, and most runs still give the key.

\begin{table}[h]
\centering\small
\caption{Analysis effort per run (Claude models).}
\label{tab:effort}
\begin{tabular}{lcccc}
\toprule
 & original & null & withdrawn & flipped \\
\midrule
agent turns (median) & 5 & 5 & 6 & 6 \\
wall time, s (median) & 30 & 27 & 49 & 38 \\
\bottomrule
\end{tabular}
\end{table}

\textbf{GeneBench-Pro per problem} (Table~\ref{tab:gbtasks}). Transform-consistent answers are common on every problem, including those where the grader almost never passes, so the strict grader hides responsiveness on all four problems. Removing the case replicates (Hi-C) or making the loci indistinguishable (selection) is flagged most often. Permuted genotypes or phenotypes (eQTL and QTL) are never flagged, and agents fit a model and report a null estimate without saying that the signal is gone. Transform consistency (53\%) is similar to the agreement between repeated runs on unchanged data (47\%), suggesting that run-to-run variability contributes to the inconsistent responses.

GPT-5.6-Sol shows the same pattern (50\% against 50\%). It follows the founder relabelling in every run and always picks the swapped locus, but its estimate of the selection coefficient varies between 0.04 and 0.10 across runs on identical data, so the selection problem reaches neither replicate agreement nor transform consistency.

Biomni shows the same pattern. The grader passes one of its four original answers (multi-parent QTL) and none of its flipped answers. Its flipped answers are transform-consistent on two problems: on Hi-C it exchanges the case and control estimates exactly, and on multi-parent QTL it moves the high-effect founder from F5 to F2 at an unchanged position, exactly as the relabelling requires. On eQTL the effect changes sign after allele recoding ($-0.38$ to $+0.25$), and on selection it picks the swapped locus, but in both cases the magnitude differs from its original estimate, as for GPT-5.6-Sol.

\begin{table}[h]
\centering\small
\caption{GeneBench-Pro per problem (\%). Pass is the official grader against the original ground truth. The Claude models and GPT-5.6-Sol have three replicates per problem.}
\label{tab:gbtasks}
\begin{tabular}{lccccc}
\toprule
 & \multicolumn{4}{c}{Claude models} & GPT-5.6-Sol \\
\cmidrule(lr){2-5}\cmidrule(lr){6-6}
problem (flip) & pass on $E$ & $I_{\text{rep}}$ & $R^{\text{flag}}_{\text{rm}}$ & $R^{\text{rel}}_{\text{flip}}$ & $R^{\text{rel}}_{\text{flip}}$ \\
\midrule
Hi-C loop strength (case$\leftrightarrow$control) & 0 & 56 & 78 & 56 & 33 \\
multi-parent QTL (founders relabelled) & 33 & 33 & 0 & 56 & 100 \\
scRNA eQTL (allele recoded) & 0 & 56 & 0 & 44 & 67 \\
Wright--Fisher selection (loci swapped) & 44 & 44 & 67 & 56 & 0 \\
\bottomrule
\end{tabular}

\end{table}

\textbf{Flip-following by no-data answer.} Flips are followed in 62\% of runs whose no-data answer was correct and in 57\% of the others, so in this run-level comparison, runs on questions answerable without data did not follow fewer flips. This comparison is across runs rather than models as in Sec.~\ref{sec:rq3}, and the anonymisation experiment remains the direct test of the prior (App.~\ref{app:anon}).

\textbf{Prior without background.} Removing the study background from the no-data prompt lowers accuracy only from 83\% to 78\%, so the answers come mostly from knowledge of the question itself.

\section{Gene-name anonymisation}
\label{app:anon}
\textbf{Construction.} The follow-up uses the 10 verified BAIS-SD tasks whose question or options name genes. They come from the marker, expression and condition differential-expression families. S38\_Q3 is excluded because its question names the gene only by function (``the rod opsin gene''), which becomes unanswerable once names are hidden. For each task, every gene symbol is mapped to an opaque identifier by a fixed random permutation, in \texttt{var\_names} and \texttt{.raw} of every file of the original, withdrawal and flip versions. All other gene annotation columns, such as symbols, Ensembl identifiers and biotypes, are dropped because they map back to symbols. The same mapping replaces gene names in marker-style cell labels (``CD14+ monocytes'') and in the question, options and background. Gene symbols that double as study vocabulary are kept. These are cluster identifiers (C1, C2 and so on), ``CTRL'' and ``RPE'', and in S16\_Q1 the mutated genes that name the experimental groups in the data (TET2 and DNMT3A). The reference statistic, recomputed on every anonymised file with the mapped names, supports exactly the same option as before on all 10 tasks, and the option genes no longer appear in the text. Because the anonymised files are derived from the verified versions, the interventions are identical.

\textbf{Runs.} The original, withdrawal, flip and no-data versions were run with the three Claude Code models and both GPT-5.6 models, three replicates each. The named results in Table~\ref{tab:anon} use the same tasks and replicates from the main runs.

\textbf{Residual prior.} S03\_Q3 stays answerable without data after anonymisation (80\%). The models recognise the study from its background and recall its finding that age-related macular degeneration genes are expressed in Müller glia, which hiding gene names cannot remove. Without this task, no-data accuracy falls from 85\% to 22\% for the Claude models and from 85\% to 20\% for the GPT models. Flips followed rise from 54\% to 93\% and from 48\% to 89\%, and evidence-grounded accuracy rises from 35\% to 65\% and from 13\% to 50\%.

\textbf{Withdrawal without names.} When the queried gene is deleted, agents check the three remaining option genes in the data, find that none fits, and choose the fourth, which they cannot find, by elimination. Their explanations say so explicitly (for example ``G21386 is not in the dataset, so C was chosen by elimination''). This is why part of the withdrawal credit survives anonymisation.

\begin{table}[h]
\centering\scriptsize
\setlength{\tabcolsep}{3pt}
\caption{Named and anonymised gene tasks (\%, 10 tasks and 3 replicates per agent). For the Claude models combined, the 95\% task-bootstrap interval of the difference excludes zero for no-data accuracy, flips followed, credit after withdrawal and after a flip, withdrawal flagged and EGA, and includes zero for accuracy on the original data.}
\label{tab:anon}
\resizebox{\linewidth}{!}{%
\begin{tabular}{lcccccccccccccc}
\toprule
 & \multicolumn{2}{c}{accuracy on $E$} & \multicolumn{2}{c}{no data} & \multicolumn{2}{c}{withdrawal flagged} & \multicolumn{2}{c}{credit after withdrawal} & \multicolumn{2}{c}{flips followed} & \multicolumn{2}{c}{credit after flip} & \multicolumn{2}{c}{EGA} \\
\cmidrule(lr){2-3}\cmidrule(lr){4-5}\cmidrule(lr){6-7}\cmidrule(lr){8-9}\cmidrule(lr){10-11}\cmidrule(lr){12-13}\cmidrule(lr){14-15}
agent & named & anon. & named & anon. & named & anon. & named & anon. & named & anon. & named & anon. & named & anon. \\
\midrule
Opus 5.5 & 97 & 100 & 90 & 40 & 40 & 40 & 83 & 73 & 60 & 100 & 37 & 0 & 37 & 60 \\
Sonnet 5 & 100 & 90 & 97 & 37 & 17 & 53 & 90 & 70 & 43 & 97 & 53 & 3 & 23 & 67 \\
Haiku 4.5 & 87 & 83 & 73 & 3 & 47 & 40 & 63 & 13 & 70 & 83 & 17 & 10 & 57 & 80 \\
GPT-5.6-Sol & 100 & 93 & 87 & 30 & 20 & 17 & 90 & 50 & 53 & 93 & 43 & 7 & 20 & 57 \\
GPT-5.6-Terra & 90 & 83 & 87 & 27 & 27 & 30 & 83 & 50 & 53 & 87 & 47 & 13 & 23 & 53 \\
Claude (all) & 94 & 91 & 87 & 27 & 34 & 44 & 79 & 52 & 58 & 93 & 36 & 4 & 39 & 69 \\
\bottomrule
\end{tabular}%
}
\end{table}

\section{Repairs}
\label{app:repairs}
The abstention experiment uses the same tasks, models and protocol, with 486 further Claude Code runs (18 tasks, 3 models, 3 replicates and 3 data versions). The grounding-focused judge regrades existing runs.

\textbf{Abstention option.} Each question receives an extra option, ``The provided data are insufficient to determine the answer'', on the original, withdrawal and flip versions (Table~\ref{tab:followup}). Agents never abstain on intact data, where accuracy is 97\% in these runs, against 95\% in the main runs without the option. They abstain in 57\% of withdrawal runs and in only 6\% of flip runs. Credit for the key falls from 77\% to 31\% on withdrawn evidence and from 30\% to 21\% on flipped evidence. Marker-gene questions remain the weakest, since agents still answer them by elimination.

\textbf{Grounding-focused judge.} The judge model and trajectories are unchanged, and the prompt adds that matching the reference is not sufficient and that answers not supported by the agent's own executed analysis must score low. Over all Claude runs, withdrawal sensitivity rises from 0.25 to 0.44 at unchanged null sensitivity (0.06 and 0.06), and the flip AUC rises from 0.11 to 0.82, against 0.12 for the benchmark score. On the common run set of Table~\ref{tab:judges}, the flip AUC rises from 0.12 to 0.83, against 0.11 for the benchmark score.

\begin{table}[h]
\centering\small
\caption{Abstention option (\%, Claude models), with the option ``The provided data are insufficient to determine the answer'' offered or not. Key credit is the official BAIS score against the original key.}
\label{tab:followup}
\begin{tabular}{lccccccc}
\toprule
 & \multicolumn{3}{c}{chooses abstention option} & \multicolumn{2}{c}{key credit, withdrawn} & \multicolumn{2}{c}{key credit, flipped} \\
\cmidrule(lr){2-4}\cmidrule(lr){5-6}\cmidrule(lr){7-8}
Model & orig & withdrawn & flipped & without & with & without & with \\
\midrule
Opus 5.5 & 0 & 59 & 6 & 85 & 37 & 33 & 19 \\
Sonnet 5 & 0 & 46 & 6 & 89 & 48 & 41 & 35 \\
Haiku 4.5 & 0 & 65 & 6 & 56 & 9 & 15 & 9 \\
all & 0 & 57 & 6 & 77 & 31 & 30 & 21 \\
\bottomrule
\end{tabular}
\end{table}

\section{Case studies}
\label{app:cases}
Table~\ref{tab:cases} shows representative runs. Every quotation is copied from the agent's own \texttt{answer.json}, and we checked automatically that each fragment occurs verbatim. ``Judge'' gives the Opus judge's overall score (0--10) under the standard and the grounding prompt.

\textbf{Answer from memory, rewarded.} After TMPRSS2 is deleted, Sonnet reports that the gene is absent and that its answer is ``a low-confidence extrapolation from \ldots\ prior knowledge''. The benchmark gives full credit, and the standard judge 6 of 10. Offered the option ``the provided data are insufficient'', the same model on the same data abstains. The information needed to abstain was already in the trajectory, but the answer format had no place for it.

\textbf{The same failure occurs across scaffolds.} When the deep-caries cells are removed, Claude Code (Opus), Codex (GPT-5.6-Sol) and Biomni (Sonnet 5) each state that no deep-caries sample is present, and each gives the key VCAM1 from its known function. All three receive full credit from the benchmark and 7 of 10 from the standard judge, while the grounding judge gives 3 of 10.

\textbf{Elimination and overruling the data.} After a flip, Opus finds that HES1 is barely expressed in matrix fibroblasts and chooses it anyway ``by ruling those out, not because the data show it''. Sonnet finds RHO highest in cones, concludes that ``the RHO column looks mis-annotated'', and answers rods ``from known biology, not from the RHO data''. Both are graded fully correct. Treating a surprising measurement as an error can be good practice, but here it is applied selectively, to exactly the evidence that contradicts the expected answer, and the benchmark rewards it.

\textbf{Following the evidence, scored zero.} On the Hi-C problem, Opus's case and control loop strengths on the flipped data are its original estimates exchanged to the third decimal, and its $\Delta$ changes sign exactly as the ground truth does. Its estimate is outside the grader's tolerance on both versions, so it scores 0 on both, the same as an agent that ignored the swap.

\begin{table}[h]
\centering\scriptsize
\setlength{\tabcolsep}{3pt}
\caption{Case studies. ``ans.'' is the agent's answer, ``bench.'' the official score against the original key, and ``judge'' the Opus judge's overall score under the standard and the grounding prompt.}
\label{tab:cases}
\resizebox{\linewidth}{!}{%
\begin{tabular}{p{2.1cm}p{2.3cm}p{1.6cm}c p{5.2cm}cc}
\toprule
case & data & run & ans. & agent's evidence (verbatim) & bench. & judge \\
\midrule
Answer from memory after noticing the gap & withdraw: TMPRSS2 deleted from every file & S00\_Q4 Sonnet 5 & D & ``TMPRSS2 is absent from the provided nasal h5ad (X and raw), so it could not be quantified.'' \ldots\ ``the answer is a low-confidence extrapolation from that plus prior knowledge.'' & 1 (key) & 6/2 \\
\midrule
Same data, abstention offered & withdraw (as above), option ``insufficient data'' added & S00\_Q4 Sonnet 5 & E & ``TMPRSS2 is absent from the nasal (and bronchial) matrices, both in X and raw'' \ldots\ ``so its expression by cell type cannot be measured.'' & abstained & -- \\
\midrule
Same failure in three scaffolds & withdraw: the deep-caries (pulpitis) cells removed & S21\_Q2 Opus 5.5 & B & ``it has no deep caries cells, so upregulation in deep caries can't be tested.'' \ldots\ ``the answer rests on VCAM1's known role in endothelial activation and leukocyte adhesion.'' & 1 (key) & 7/3 \\
 &  & S21\_Q2 GPT-5.6-Sol & B & ``the h5ad contains only normal (890 cells) and enamel-caries (2,015 cells) samples and no deep-caries sample'' & 1 (key) & 7/3 \\
 &  & S21\_Q2 Biomni (Sonnet 5) & B & ``The provided h5ad file contains only 'normal' and 'enamel caries' samples (not 'deep dental caries' as described in the background)'' & 1 (key) & 7/3 \\
\midrule
Elimination instead of evidence & flip: HES1 and PTPRC expression swapped & S35\_Q2 Opus 5.5 & B & ``they barely express HES1 (mean 0.02, 1\% of cells).'' \ldots\ ``HES1 was chosen by ruling those out, not because the data show it.'' & 1 (key) & 6/3 \\
\midrule
The data are declared wrong & flip: RHO expression moved from rods to cones & S38\_Q3 Sonnet 5 & D & ``the feature labelled RHO (ENSG00000163914) is highest in cones ($\sim$55 vs $\sim$0.8 in rods), which contradicts the rod answer.'' \ldots\ ``so the labels are right and the RHO column looks mis-annotated. I chose Rods (D) from known biology, not from the RHO data.'' & 1 (key) & 6/2 \\
\midrule
Following the evidence, scored zero & flip: case and control replicates swapped (Hi-C) & Hi-C Opus 5.5 & -- & $\Delta$ = 1.87 on $E$, -1.87 on $T_CE$ (ground truth 2.40 and -2.40), with the case and control estimates exchanged exactly & 0 on both & 2/3 \\
\bottomrule
\end{tabular}%
}
\end{table}

\end{document}